# Segmentation and Classification of Pap Smear Images for Cervical Cancer Detection Using Deep Learning

**Nisreen Albzour and Sarah S. Lam**
*Binghamton University, Binghamton, NY, USA*

## Abstract

Cervical cancer remains a significant global health concern and a leading cause of cancer-related deaths among women. Early detection through Pap smear tests is essential to reduce mortality rates; however, the manual examination is time-consuming and prone to human error. This study proposes a deep learning framework that integrates U-Net for segmentation and a classification model to enhance diagnostic performance. The Herlev Pap Smear Dataset, a publicly available cervical cell dataset, was utilized for training and evaluation. The impact of segmentation on classification performance was evaluated by comparing the model trained on segmented images and another trained on non-segmented images. Experimental results showed that the use of segmented images marginally improved the model's performance on precision (+0.41%) and F1-score (+1.30%), which suggests a slightly more balanced classification performance. While segmentation helps in feature extraction, this research's results showed that its impact on classification performance appears to be limited. The proposed framework offers a supplemental tool for clinical applications, which aids pathologists in early diagnosis.

## 1. Introduction
Cervical cancer remains a significant global health challenge, in which over 600,000 new cases and 340,000 deaths are reported annually [1]. The primary strategy to reduce mortality is early detection through Pap smear tests, which enable the identification of precancerous lesions. However, manual analysis of Pap smear slides is time-consuming, prone to human error, and subject to inter-observer variability, which highlights the need for automated diagnostic solutions [2, 3].

Deep Learning (DL), particularly Convolutional Neural Networks (CNNs), has revolutionized medical image analysis by providing efficient and accurate diagnostic tools [4, 5]. CNNs have demonstrated remarkable success in detecting cancerous patterns within medical images, which surpass traditional machine learning methods by automatic extraction of hierarchical features [6, 7]. Automated and hyperautomation-based AI pipelines have further extended these capabilities to other medical-imaging domains, such as hematologic malignancy detection [8, 9]. Additionally, preprocessing techniques such as segmentation enhance CNN performance by isolating diagnostically relevant regions and minimizing background noise [10, 11]. Studies have also shown that data augmentation techniques—such as rotation, flipping, and scaling—help improve model robustness and address class imbalance, which further enhance classification accuracy [12, 13]. Moreover, advanced feature selection methods such as metaheuristic algorithms optimize classification performance, which ensures the model focuses on the most relevant information [14]; such feature-selection strategies have likewise improved predictive performance in other clinical machine-learning applications, including post-stroke rehabilitation outcome prediction [15].

This study employs the Herlev Pap Smear Dataset to develop a binary classification framework that distinguishes between normal and abnormal cervical cells. To refine segmentation quality, preprocessing techniques—including adaptive thresholding, Gaussian blurring, and morphological transformations—are applied. Furthermore, Class Activation Maps (CAMs) improve model interpretability by highlighting critical regions that influence classification decisions [16]. Explainable AI methods are increasingly being integrated into deep learning frameworks to enhance transparency and clinical trust in automated diagnostic systems [17], and recent work has examined how calibration and explainability transfer across cohorts under distribution shift [18].

Recent related work has expanded automated clinical machine-learning pipelines in several directions. In cervical cytology specifically, transformer-based studies have shown that Vision Transformer architectures can provide competitive classification performance, statistical validation, deployment efficiency, and improved interpretability for Pap smear image analysis [19, 20]. These developments suggest that the segmentation-based CNN framework evaluated in this study can serve as a useful baseline for future comparisons with transformer-based and reliability-aware diagnostic systems.

By leveraging state-of-the-art deep learning methodologies, this research aims to develop a scalable and accurate diagnostic framework for automated cervical cancer screening. The findings contribute to improving early detection efforts and bridging existing gaps in AI-driven medical diagnostics.

## 2. Methodology

Figure 1 provides an overview of the entire methodology employed in this study, which covers various key stages such as dataset preparation, image segmentation, data augmentation, the design of the CNN architecture, and model evaluation. Each stage plays a crucial role in ensuring the effectiveness and robustness of the classification framework, which underscores the systematic approach adopted to tackle the challenges in cervical cancer diagnosis.

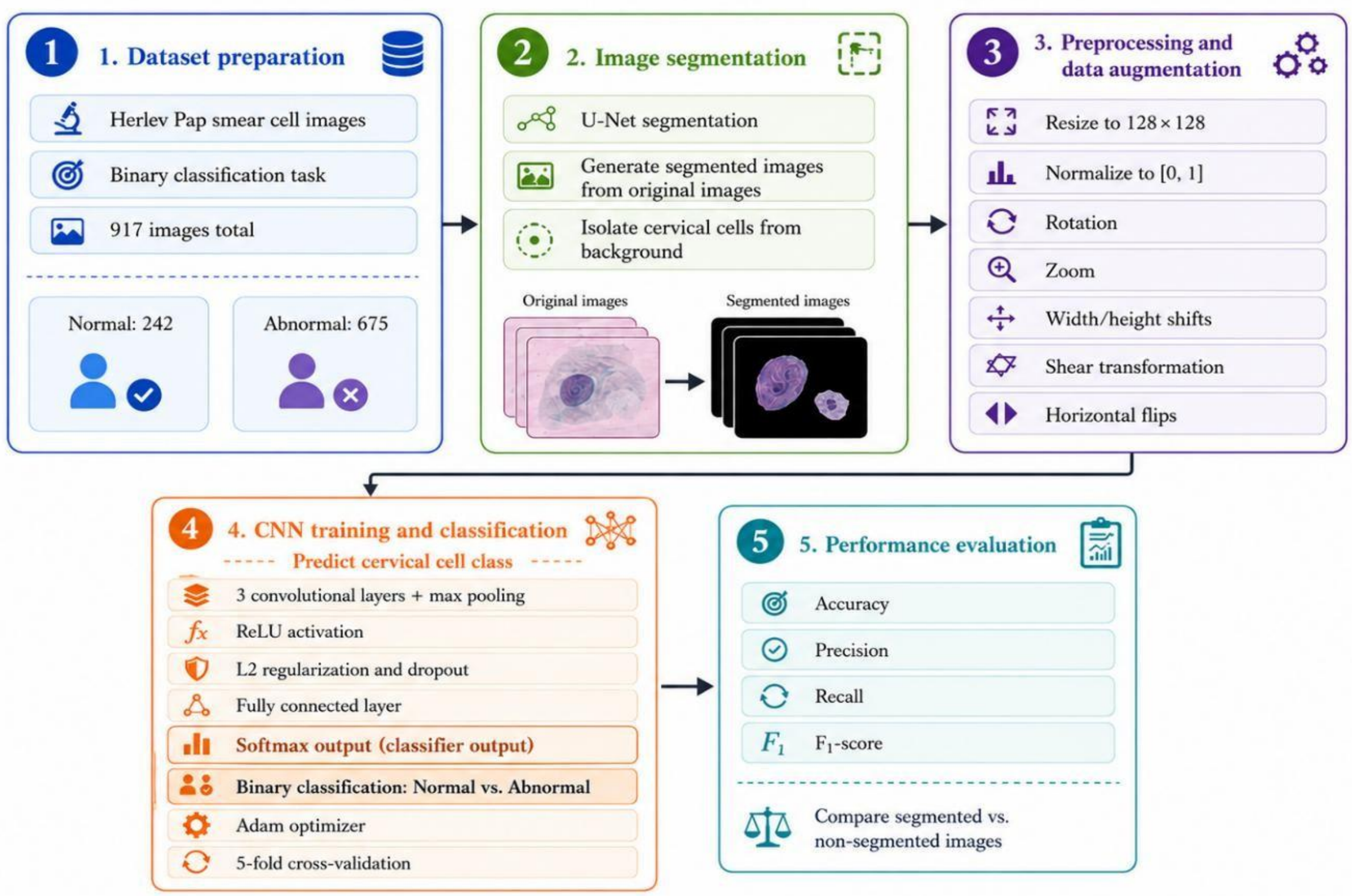

Figure 1: Overview of the Methodology

### 2.1. Dataset Description

The Herlev Pap Smear Dataset, a benchmark resource for cervical cancer research, is a widely used collection of cervical cell images. Originally, the dataset contained 917 images classified into seven distinct categories based on the stages of cervical cell abnormalities and normal conditions. These categories were designed to address multiclass classification problems, which focus on a more granular analysis of cervical cell types. However, for the purposes of this study, the dataset was restructured into a binary classification format to simplify the analysis and improve the applicability of the results in clinical diagnostics. The images were grouped into two broad categories:

- **Normal**: This class includes images that belong to normal epithelial cells, which represent healthy cervical conditions.
- **Abnormal**: This class encompasses all dysplastic and carcinoma cell classes, which represent various stages of cervical abnormalities and malignancies.

The restructured dataset, detailed in Table 1, comprises 242 images labeled as "Normal" and 675 images labeled as "Abnormal." To ensure a comprehensive evaluation of the model's generalization capability, 5-fold cross-validation was implemented instead of a training/testing split. In this approach, the dataset was divided into five subsets, where a model was trained on four subsets and tested on the left-out subset in an iterative manner. This ensures that all 917 images were utilized for both training and validation across different folds. Each image was resized to a resolution of 128×128 pixels to ensure consistency in input dimensions while preserving the visual characteristics necessary for classification. This preprocessing step maintains compatibility with the model architecture while retaining key morphological features for accurate diagnosis.

Table 1: Class Distribution

| Class | Total |
|---|---|
| Normal | 242 |
| Abnormal | 675 |
| **Total** | **917** |

### 2.2. Image Segmentation

To enhance cervical cancer classification, image segmentation was performed using a U-Net model, a widely used deep learning architecture for biomedical image segmentation. The Herlev Dataset images were first preprocessed by converting them to grayscale, resizing them to 128×128 pixels, and normalizing pixel values to [0, 1]. The pretrained U-Net model was then applied to generate segmentation masks, which isolates cervical cells from the background. Each image was fed into the model, which produced a corresponding segmented mask, which highlights the cell regions while reducing noise.

The segmented masks were post-processed by rescaling pixel values to [0, 255]. This segmentation step helped improve feature extraction, which ensures that only relevant regions were used in the subsequent classification phase. Figure 2 illustrates the impact of the U-Net segmentation model on Pap smear images. Figure 2(a) depicts the original image before segmentation, while Figure 2(b) shows the corresponding segmented image, which highlights cervical cell boundaries. The segmentation process effectively isolates cells from the background, which enhances feature extraction for classification.

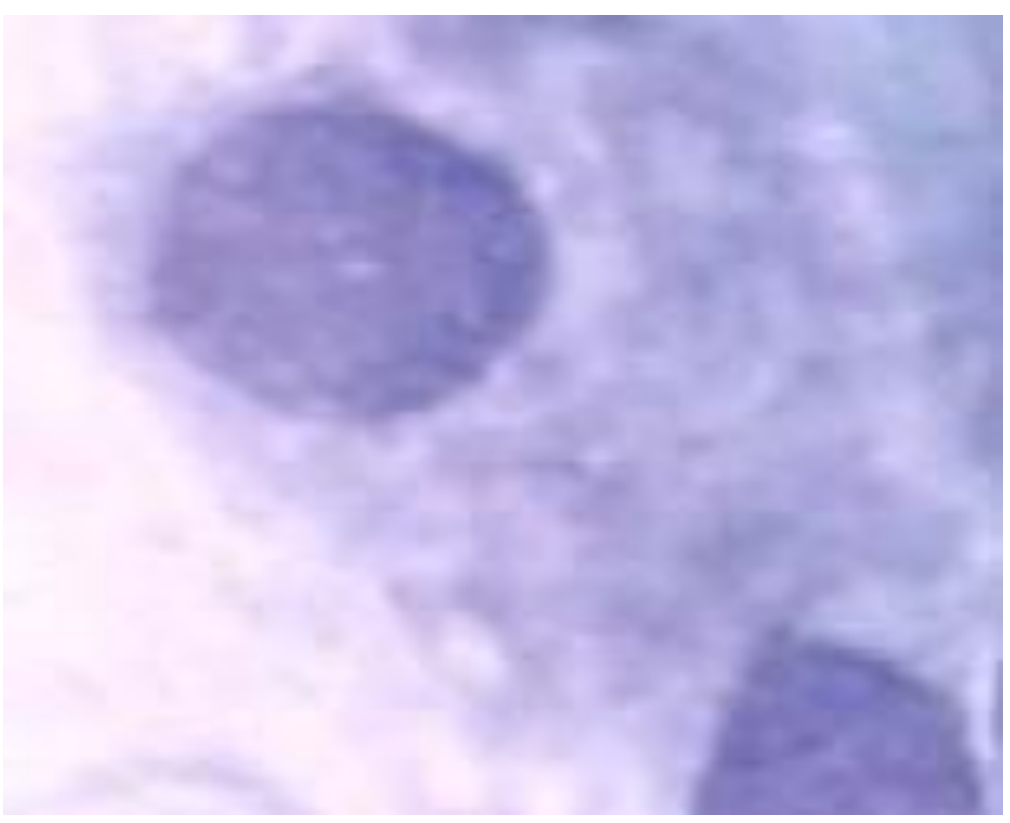

(a) Before segmentation

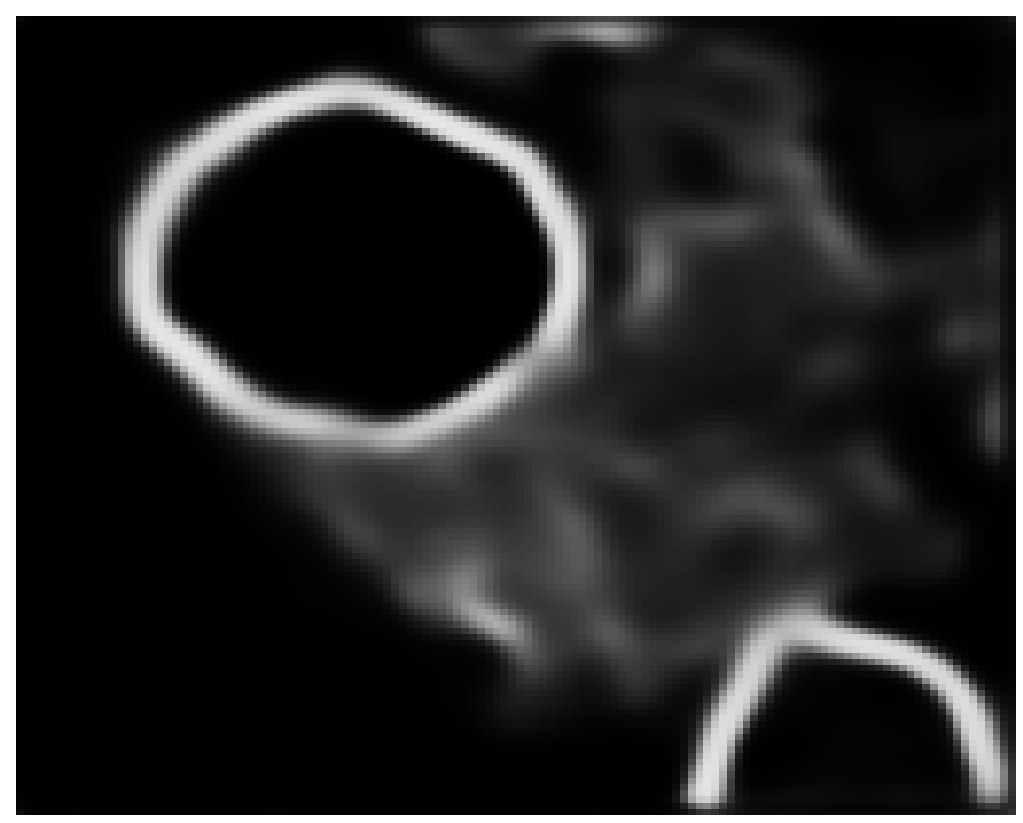

(b) After segmentation

Figure 2: Comparison of Segmentation Results

### 2.3. Data Augmentation

To enhance the robustness and generalization of the classification model, data augmentation was employed during preprocessing. Given the limited availability of cervical cell images, augmentation techniques such as flipping, rotation, and contrast adjustments were applied to increase dataset diversity. This helped prevent overfitting and improved the model's ability to generalize to unseen samples. Additionally, images were normalized to a [0, 1] range to ensure consistency in pixel intensity and resized to a fixed dimension of 128×128 pixels before being used in training.

### 2.4. CNN Architecture for Classification

CNN was designed to distinguish between normal and abnormal cervical cells (i.e., a classification task). The model consisted of three convolutional layers with increasing filter sizes (32, 64, and 128), each followed by max pooling to reduce spatial dimensions. ReLU activation was used to introduce non-linearity, L2 regularization and dropout were included to help mitigate overfitting. A fully connected layer with 128 neurons was followed by a softmax output layer that performed binary classification. The model was trained using categorical cross-entropy loss and optimized with the Adam optimizer. To evaluate performance, 5-fold cross-validation was conducted on both segmented and non-segmented images, in which accuracy, precision, recall, and F1-score were used as performance metrics. Clinical AI models may be affected by data imbalance, feature-selection choices, model-selection bias, and limited generalizability. Beyond model-level concerns, prior work has also examined how medical software can produce unintended effects on clinical decisions and patient safety, underscoring the importance of evaluating such systems from a broader sociotechnical perspective [21]. Related clinical machine-learning studies have emphasized the importance of robust evaluation protocols and statistical validation across different healthcare applications, including breast cancer survival prediction [22]. Although the present study focuses on a compact CNN-based comparison between segmented and non-segmented Pap smear images, the use of cross-validation provides a foundation for future extensions involving more rigorous statistical testing, external validation, and model reliability assessment.

## 3. Results and Discussion

### 3.1. Non-Segmented Data Performance

The classification results for non-segmented images are summarized in Figure 3, which presents the confusion matrix. The model demonstrates moderate performance in distinguishing between normal and abnormal cervical cells, which achieves an accuracy of 81.02%. However, the false positive rate remains relatively high; 144 of the normal epithelial cells were misclassified as abnormal cells. This indicates that while the model trained on non-segmented images provides reasonable classification capability, there is still room for improvement in precision and recall balance. Figure 4 illustrates the performance of the validation models trained using the 5-fold cross-validation approach. The figure shows that the validation models experience fluctuations in accuracy (some level of instability) during the training process.

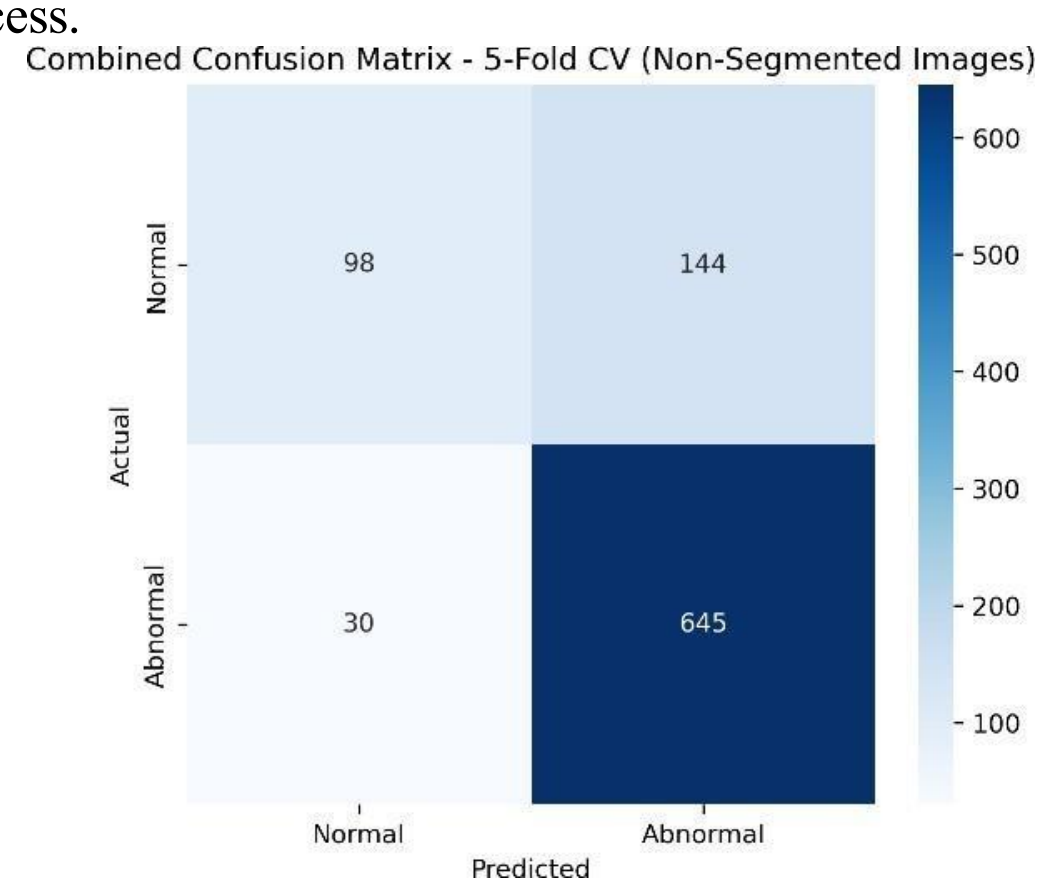

Figure 3: Confusion Matrix—Non-Segmented

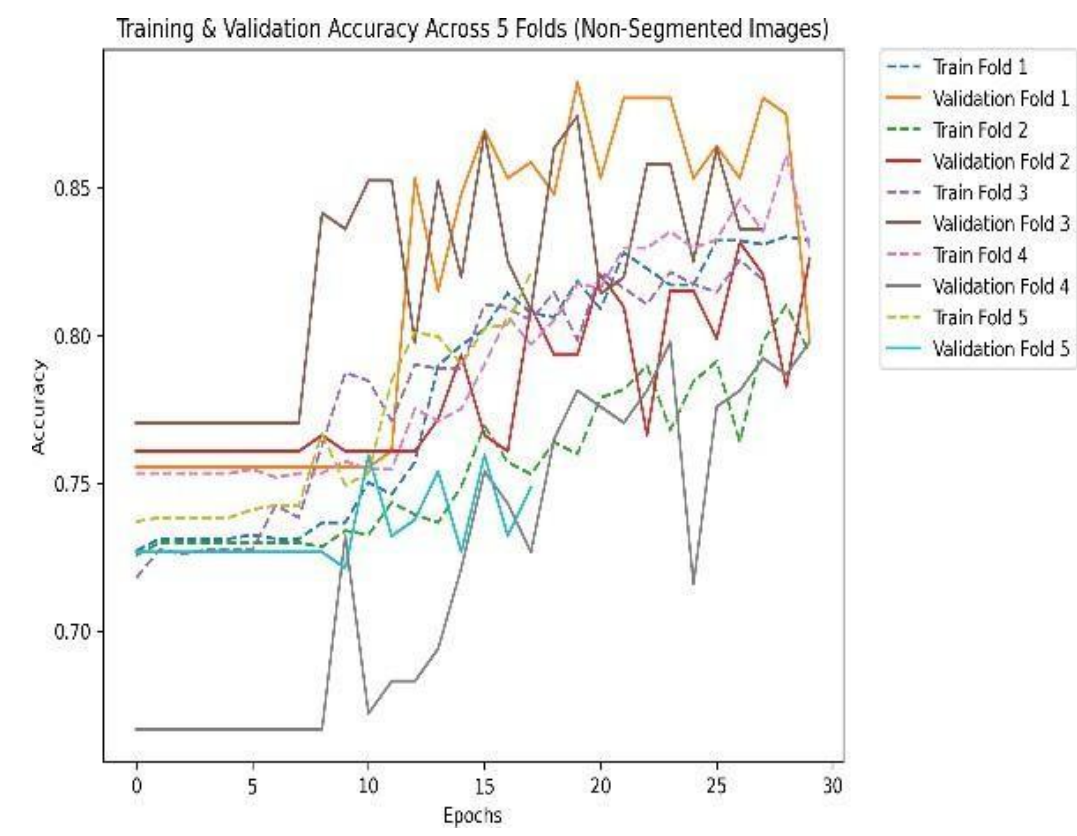

Figure 4: Training Accuracy—Non-Segmented

### 3.2. Segmented Data Performance

The classification results for segmented images are summarized in Figure 5. The model that was trained on segmented images achieves an accuracy of 80.80%, which is similar to the accuracy of the model that was trained on non-segmented images. The false positive rate remains relatively high; 123 of the normal epithelial cells were misclassified as abnormal cells. This indicates that while the model trained on segmented images provides at par classification capability, there is still room for improvement in precision and recall balance. Figure 6 illustrates the performance of the validation models trained on segmented data using the 5-fold cross-validation approach. The figure shows that the validation models experience less fluctuations in accuracy during the training process, when compared to the case for non-segmented data (as discussed in section 3.1). This may have been attributed to the segmented dataset that allows the model to learn more discriminative features.

### 3.3. Performance Comparison

A quantitative comparison between the two models' performance on segmented and non-segmented datasets is provided in Table 2. The evaluation was performed using 5-fold cross-validation.

Table 2: Performance Comparison Between Segmented and Non-Segmented Data

| **Metric** | **Non-Segmented** | **Segmented** | **Improvement** |
|---|---|---|---|
| **Accuracy** | 81.02% | 80.80% | **-0.22%** |
| **Precision** | 80.44% | 80.85% | **+0.41%** |
| **Recall** | 81.02% | 80.80% | **-0.22%** |
| **F1-score** | 78.25% | 79.55% | **+1.30%** |

The two CNN models, one trained on non-segmented images and the other trained on segmented images, resulted in similar performance across all evaluation metrics. Overall, the use of segmented images in model development resulted in an improvement of 1.30% in F1-score. This may indicate that segmentation aids in achieving a more balanced classification performance. While segmentation can help extract relevant features and reduce background noise in images, its overall contribution to classification accuracy may depend on the complexity of images and the model's design and parameter choices. The modest gain in F1-score observed with segmented images should therefore be interpreted as an early indication rather than definitive evidence that segmentation improves cervical cell classification. Recent cervical-cell studies using Vision Transformers and CNN baselines have reported stronger classification performance under more extensive optimization and statistical validation protocols [19, 20]. Therefore, the present findings are best viewed as a baseline segmentation-versus-non-segmentation comparison that motivates future benchmarking against compact transformers, ensemble models, and calibrated decision-support frameworks.

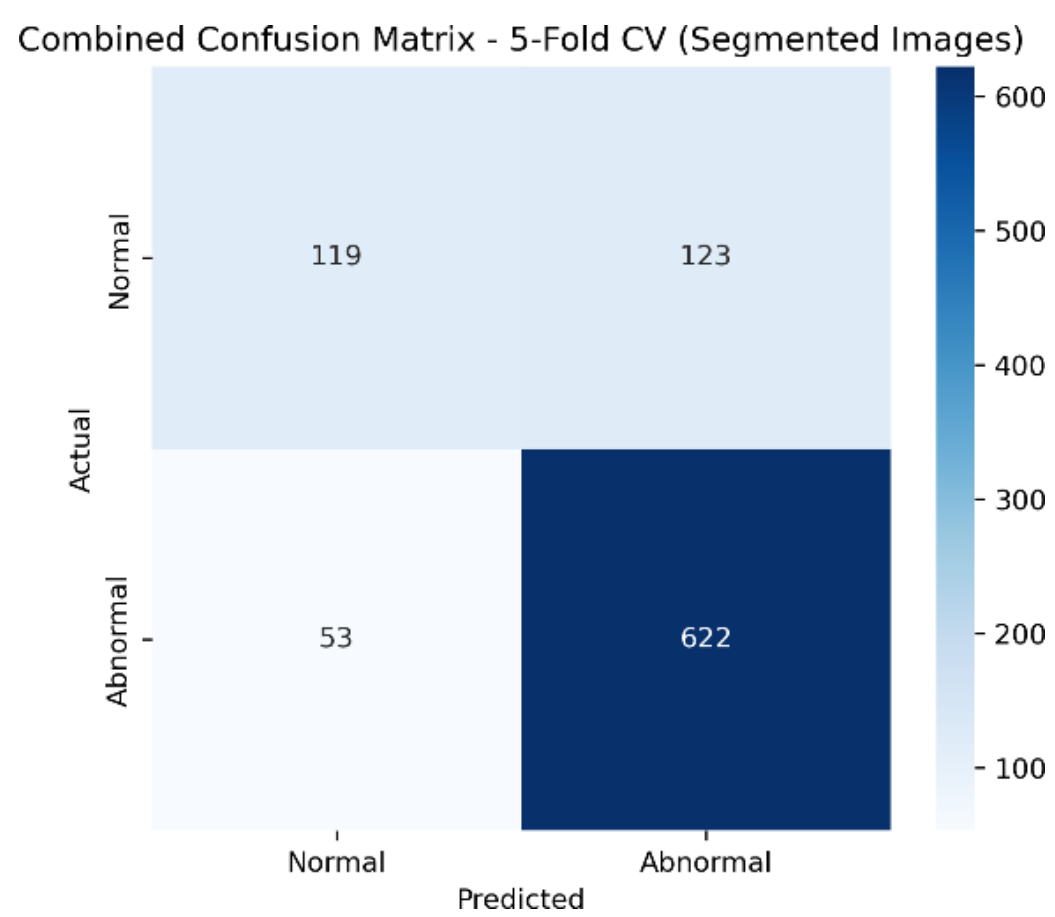

Figure 5: Confusion Matrix—Segmented

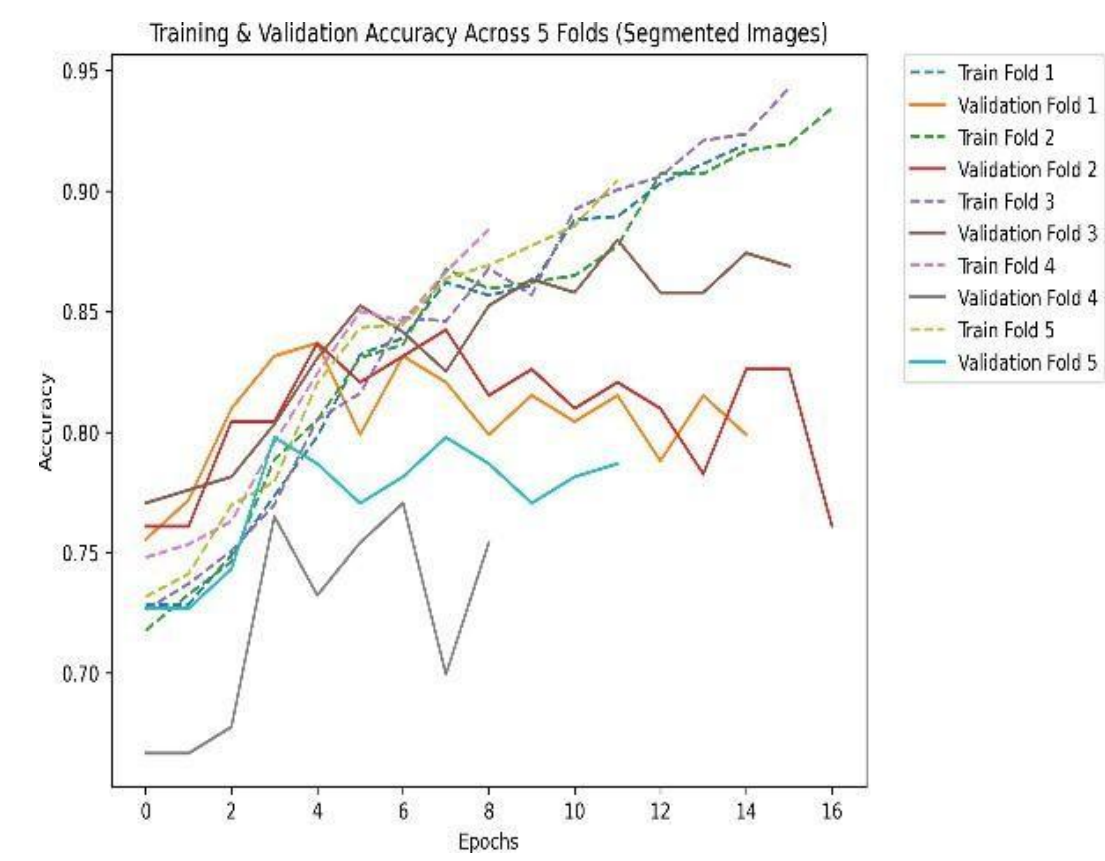

Figure 6: Training Accuracy—Segmented

## 4. Conclusion

This study evaluated the impact of image segmentation on cervical cancer classification using Pap smear images. A U-Net model was used to segment cervical images. Two CNNs were developed for classification of cervical cancer; one trained on non-segmented images and another trained on segmented images. Five-fold cross-validation was used in both cases. The results for both CNNs are comparable in terms of accuracy, precision, and recall. The model that was trained on segmented images shows a marginal improvement of 1.30% in F1-score, when compared to the model trained on non-segmented images, which suggests better balance between precision and recall. While segmentation helps extract relevant features, its impact on classification performance appears to be limited.

Future work could explore enhanced preprocessing techniques, adaptive feature selection, and advanced deep learning models to further refine classification accuracy and model reliability. In particular, future extensions should compare the present CNN-based segmentation framework with compact Vision Transformers, hybrid/ensemble architectures, and calibration-aware evaluation pipelines. Recent work in clinical machine learning has shown that model complexity does not always translate into robust generalization under leakage-free and external validation protocols [9], while calibration and explainability may transfer differently across cohorts under distribution shift [18]. These findings reinforce the need to evaluate future cervical cancer screening models not only by accuracy, precision, recall, and F1-score, but also by statistical robustness, calibration, explainability, and deployment efficiency.